\documentclass[11pt]{article}
\usepackage{float} 
\usepackage{hyperref}
\usepackage{amsfonts}
\usepackage{amsbsy}
\usepackage{amsmath}
\usepackage{amssymb}
\usepackage{amsthm}
\usepackage{rotating}
\usepackage{graphics}
\usepackage{epstopdf}
\usepackage[numbers]{natbib}
\usepackage{caption}
\usepackage{subcaption}
\usepackage{xcolor}
\usepackage{bm,bbm,lscape}
\usepackage{booktabs}
\usepackage{longtable}
\usepackage{enumitem}
\usepackage{listings}
\usepackage{adjustbox}
\usepackage{algorithm}
\usepackage{resizegather}

\usepackage{ulem}

\usepackage[
top    = 1in,
bottom = 1in,
left   = 1in,
right  = 1in]{geometry}

\usepackage{setspace} 
\graphicspath{{../}}

\newtheorem{theorem}{Theorem}

\newtheorem{corollary}{Corollary}

\newtheorem{property}{Property}
\newcommand{\distas}[1]{\mathbin{\overset{#1}{\kern\z@\sim}}}%

\def\rhoo{{\rho_*(\bm{l}|\bm{x})}}

\DeclareMathOperator*{\argmin}{argmin}

\def\hat{\widehat}
\def\tilde{\widetilde}

\def\cal{\mathcal}
\def\calS{{\cal S}} 

\def\calD{{\cal D}} 

\begin{document}
\title
{\bf Query-augmented Active Metric Learning}
\author
{
Yujia Deng\footnote{These authors contributed equally to this work}\\
Department of Statistics, University of Illinois, Urbana-Champaign,\\
Yubai Yuan\footnote{These authors contributed equally to this work}\\
Department of Statistics, University of California, Irvine,\\
Haoda Fu\\
Eli Lilly and Company,\\
and\\
Annie Qu\\
Department of Statistics, University of California, Irvine\\
}
\date{}
\maketitle
\begin{abstract}

In this paper we propose an active metric learning method for clustering with pairwise constraints. The proposed method actively queries the label of informative instance pairs, while estimating underlying metrics by incorporating unlabeled instance pairs, which leads to a more accurate and efficient clustering process. In particular, we augment the queried constraints by generating more pairwise labels to provide additional information in learning a metric to enhance clustering performance. {Furthermore, we increase the {robustness} of metric learning by updating the learned metric sequentially and penalizing the irrelevant features adaptively.} In addition, we propose a novel active query strategy that evaluates the information gain of instance pairs more accurately by incorporating the neighborhood structure, which improves clustering efficiency without extra labeling cost. In theory, we provide a tighter error bound of the proposed metric learning method utilizing augmented queries compared with methods using existing constraints only. We also investigate the improvement using the active query strategy instead of random selection. Numerical studies on simulation settings and real datasets indicate that the proposed method is especially advantageous when the signal-to-noise ratio between significant features and irrelevant features is low.
\color{black}
\end{abstract}
\color{black}

\newpage
\section{Introduction}
In recent years active learning has become a popular subfield of machine learning since the performance of any supervised learning system fundamentally relies on labeled instances which are difficult or expensive to obtain in many applications. For example, the rise of electronic medical records introduces huge amounts of medical data which could be overwhelming and infeasible for the doctors to examine for the entire population. 
Instead of labeling the dataset instance by instance, a more efficient way is to cluster the data first and let the machine auto-label the dataset based on the similarity between the unlabeled instances and a few labeled representatives from each cluster. Ideally, the clustering criteria should be able to adjust sequentially through feedback from experts. Meanwhile, the auto-labeling process is expected to query the pivotal instances actively to accelerate model training and thus to reduce costs.


The idea of incorporating experts' domain knowledge or user's feedback has been pursued in previous clustering methods {\cite{Wagstaff2001ConstrainedKC,basu_active_2004,basu_probabilistic_2006,davidson2006measuring,lu_semi-supervised_2007,liu_fast_2017}}. Specifically, a user can specify that two instances must either belong to the same cluster or two different clusters. Then the clustering procedure selects the optimal label assignment by penalizing the assignments that violate these pairwise constraints. Alternatively, instead of directly clustering the instances in the original feature space, metric learning approaches { \cite{xing_distance_2003,niu_seraph:_2011,yang_bayesian_2012,hoi_semi-supervised_2010}} seek a specific distance metric trained from the constraints. The essential goal of metric learning is to identify an appropriate distance metric that encourages “similar” objects to be close together while separating “dissimilar” objects, which improves the performance of the subsequent clustering process.

However, the aforementioned metric learning process could be inefficient and unstable since randomly-chosen constraints may provide little information about the cluster structure. To solve this problem, several active learning solutions have been proposed
For example, {\cite{basu_active_2004} proposed a \textit{explore-consolidate} framework which seeks the skeleton points that are dissimilar to each other first and then use  similar pairs to refine the boundary of the clusters. This method is then generalized by \cite{mallapragada_active_2008} in selecting the most important pairs in the consolidate phase.} \cite{grira_active_2005} proposed active fuzzy constrained clustering, which sequentially queries and collects the labels for instances under current boundary of clusters.
 Alternatively, { \cite{huang_text_2006,xiong_active_2014,biswas_active_2014} proposed different models to quantify the uncertainty of the unlabeled pairs. Other methods include \cite{mai_active_2013} using neighborhood information in density-based clustering, \cite{kok_constraint_2007} building an ensemble framework for unlabeled pair selection, and \cite{van_craenendonck_cobra:_2018} utilizing the propagation of similarity relations.}

{However, approaches exploring active clustering methods which incorporate metric learning simultaneously are still limited.} To learn the latent metric from pairwise constraints, \cite{yang_bayesian_2012} proposed an active Bayesian metric learning that defines a Mahalanobis distance between instances and actively queries using entropy based criteria; \cite{xiong_active_2014} proposed an instance-level uncertainty-based active query strategy combining metric pairwise constrained Kmeans (MPCKmeans \cite{bilenko_integrating_2004}). 
The main drawback of their methods is that they do not utilize the unlabeled instance pairs in learning the metric. Although the number of pairwise constraints provided by the user is limited, the relationships between the unlabeled instances can still be inferred based on the clustering structure, which could supply additional information and therefore improve the learning efficiency.
Another limitation on the existing active clustering methods is that they do not  utilize a dimension reduction strategy for raw data during human-machine interaction. However, identifying and selecting  significant features which are consistent with a user's clustering principles 
are very  important for enhancing the similarity within a cluster, and to achieve a more robust and consistent clustering outcome. In addition, dimension reduction also leads to more interpretable clustering criteria from experts.
Furthermore, existing models are typically retrained each time the new constraints are added, while the history of  training results is ignored. This results in a loss of information which could be utilized to improve clustering performance.

In this paper, we propose a new active clustering method with query augmentation and metric aggregation. The novelty of the proposed method is that we incorporate both pairwise constraints from the user's feedback and the implicit constraints inferred based on the clustering structure to learn the metric. We integrate the unlabeled instance pairs into the metric learning process through augmented constraints weighted by uncertainty measurement, which leads to more efficient recovery of the underlying feature space. Another novelty is that we pursue dimension reduction by penalizing the irrelevant features adaptively, based on the history of metric learning results in the sequential querying process. Thus we obtain more precise and robust clustering results consistent with the user's feedback. In addition, we propose a new query strategy based on the expected entropy change. Compared with existing active learning methods, we can incorporate the neighborhood structure and transitivity of the constraints through uncertainty measurement, which provides a more accurate evaluation of the potential effect from the queried constraints. Theoretical and numerical results confirm that the proposed method improves clustering accuracy without adding labeling cost.

The paper is organized as follows. Section \ref{sec:notation_and_background} introduces notations and background for metric learning and active-semi-supervised clustering. Section \ref{sec:methodology} presents a new active metric learning framework and the metric aggregation method. Section \ref{sec:implementation} introduces a new algorithm to implement the proposed method. Section \ref{sec:theory} establishes the theoretical results. Section \ref{sec:simulations} provides the simulation properties of the proposed active learning. Section \ref{sec:real_data} illustrates the application of the proposed method for three real datasets. The last section provides concluding remarks and discussion.

\color{black}
\section{Notation and Background} 
\label{sec:notation_and_background}
Given $n$ data points in a $p$-dimensional feature space, i.e., $\bm{x}_i \in \mathbb{R}^p, i=1,...,n$, we assume each $\bm{x}_i$ is sampled from one of the $K$ clusters and denote the cluster membership vector as $\bm{l}=(\ell_1,\cdots,\ell_n)$, where $\ell_i\in \{1,\cdots, K\}$. {For the identifiability of the cluster label, we define $\bm{l}^{(1)} = \bm{l}^{(2)}$ if there is a permutation map $\Gamma$ of $\{1, \ldots, K\}$ such that $\ell_i^{(1)}=\Gamma(\ell_i^{(2)}),~i=1, ..., n$. }Let the sample space of $\bm{l}$ be $\Omega$, and then the cardinality $|\Omega|$ equals the total number of ways to partition a set of $n$ objects into $K$ non-empty subsets {up to label switching}.
We also denote the similarity matrix as $Y\in \mathbb{R}^{n\times n}$, where $y_{ij}=1$ if $\bm{x}_i$ and $\bm{x}_j$ are in the same cluster, and $0$ otherwise. Since there is a one-to-one map between $Y$ and $\bm{\ell}$ determined by $y_{ij}=\mathbbm{1}(\ell_i=\ell_j)$, the goal of clustering can be defined as the estimation of  either $Y$ or $\bm{l}$. In \textit{unsupervised} clustering, no elements of $Y$ are known beforehand, 
while in \textit{semi-supervised} clustering, a part of the elements of $Y$ are queried from users as pairwise constraints.  These pairwise constraints are referred to as ``similar'' and ``dissimilar'' pairs whose index sets are denoted as $\calS = \{(i,j) | y_{ij}=1\}$ and $\calD = \{(i, j) | y_{ij}=0\}$, respectively, while the unlabeled set is denoted as $\mathcal{U}=\left\{\left({i}, {j}\right) |\left({i}, {j}\right) \notin \mathcal{S} \cup \mathcal{D}\right\}$. The pairwise constraints have the following transitivity property:
\begin{property}[Transitivity]
For different indexes $i, j, k$, if $(i,j)\in \mathcal{S}$ and $(i,k)\in \mathcal{S}$, then $(j, k)\in \mathcal{S}$. If $(i,j)\in \mathcal{S}$ and $(i,k)\in \mathcal{D}$, then $(j, k)\in \mathcal{D}$.
\end{property}
\noindent
The transitivity property allows us to generate more constraints within one query, which is essential in improving the efficiency of a query strategy.

In addition, we consider the case where the cluster structure under a probabilistic model can be captured in a linear subspace $\mathbb{R}^r\subset \mathbb{R}^p,~r\leq p$; i.e., there exists a matrix $M\in \mathbb{R}^{r\times p}$ with orthogonal columns such that $P(y_{ij}=1|\bm{x}_i, \bm{x}_j)=P(y_{ij}=1|M\bm{x}_i, M\bm{x}_j)$. {This low-dimensional clustering structure can be captured by a Mahalanobis distance $\left\|\bm{x}_{i}-\bm{x}_{j}\right\|_{A}^2=\left(\bm{x}_{i}-\bm{x}_{j}\right)^{\top} A\left(\bm{x}_{i}-\bm{x}_{j}\right)$, where $A=M^\top M$ is called the \textit{metric matrix}. 
The raw distance metric may involve irrelevant features which are not accurate for measuring the distances between data points; however, we can improve the clustering performance by identifying $A$.} Intuitively, the distance $\left\|\bm{x}_{i}-\bm{x}_{j}\right\|_{A}$ should be small if $\bm{x}_i$ and $\bm{x}_j$ belong to the same cluster and large if they are in different clusters. Therefore, one metric learning method \cite{xing_distance_2003} to learn $A$ is through 
\begin{equation}\label{eqn:xing}
   {\min _{A}}  {\sum_{\left({i}, {j}\right) \in \mathcal{S}}\left\|\bm{x}_{i}-\bm{x}_{j}\right\|_{A}^{2}},   \quad {\text { s.t. }}  {\sum_{\left({i}, {j}\right) \in \mathcal{D}}\left\|\bm{x}_{i}-\bm{x}_{j}\right\|_{A} \geq 1},  \quad \mbox{and} \quad {A \succeq 0}, 
\end{equation}
where $A \succeq 0$ denotes that $A$ is positive semi-definite. The above training process (\ref{eqn:xing}) is able to minimize the distances between similar pairs while separating dissimilar pairs to avoid trivial solutions with all zeros.

Moreover, as we intend to \textit{actively} query pairwise constraints sequentially, we denote the pairwise constraints at the $t$th step as $\mathcal{S}^t$ and $\mathcal{D}^t,~ t=1,\cdots, T$, where $T$ is the total number of queries called the \textit{budget}. The goal of this paper is to improve metric learning efficiency  and  design a sequential query strategy to achieve better clustering performance with a given budget. 

\section{Methodology} 
\label{sec:methodology}
{In this section, we propose an efficient metric learning method with augmented pairwise constraints through introducing a selective penalty to exclude irrelevant features, and design an active query strategy based on a new uncertainty criterion.}

\subsection{Metric learning with augmented pairwise constraints} 
\label{sub:metric_learning_with_augmented_membership}
We start with how to efficiently utilize limited numbers of pairwise constraints to train metrics. One common problem of (\ref{eqn:xing}) and existing metric learning methods \cite{Wagstaff2001ConstrainedKC,lu_semi-supervised_2007,grira_active_2005} is that only the violations on queried pairwise constraints are penalized. However, these queried constraints also provide additional prior information on other unlabeled neighborhood pairwise relations implicitly through the underlying cluster structure.
To solve this problem, we generalize the queried pairwise constraints $\mathcal{S}\cup \mathcal{D}$ to all $y_{ij}$'s by inferring the labels of unlabeled instance pairs, and train the metric matrix $A$ with both the queried pairwise constraints and the inferred pairwise constraints. 

Specifically,  we first solve for a fuzzy membership matrix $H\in\mathbb{R}^{n\times K}$ by
\begin{equation}\label{eqn:impute_H}
\begin{split}
	&\hat{H} = \argmin_{H}\sum_{(i,j)\in \mathcal{S}\cup \mathcal{D}}(y_{ij} - \bm{h}_i^T\bm{h}_j)^2 + \lambda {\sum_{i=1}^n \sum_{k=1}^K \min\left(|h_{ik}|, |h_{ik}-1|\right)},\\
    &   \mbox{s.t.} \quad h_{ij}\geq 0,\quad \sum_{k=1}^K h_{ik}=1, \text{ for all }i,
\end{split}
\end{equation}
where $\bm{h}^T_i$ is the $i$th row of $H$, and ${h}_{ik}\in [0,1]$ represents the probability that the $i$th sample belongs to the $k$th cluster. {The penalty term $\min\left(|h_{ik}|, |h_{ik}-1|\right)$ is a multi-directional separation penalty (MDSP) \cite{tang2020individualized}, which penalizes $h_{ij}$ to either 0 or 1 depending on the magnitude of $h_{ij}$. The purpose of adding the MDSP penalty is to prevent strong signals from being pulled towards zero in the process of shrinking weak signals for sparsity pursuit, and thus to reduce the uncertainty on the cluster membership of each instance.} In addition, {we only infer $\bm{h}_{i}$ if at least one element of $\{y_{i\cdot}\}$'s is observed; otherwise we let all the elements of $\bm{h}_i$ be $1/K$. }Note that the augmenting process (\ref{eqn:impute_H})  uses only the queried constraint information without involving the distance between data points since the distance metric is inaccurate during training, which may lead to biased membership inference.

Next, we utilize $\hat{H}$ to introduce additional pairwise constraints through the concordance $\hat{\bm{h}}_i^\top \hat{\bm{h}}_j$. The idea is that $\bm{x}_i$ and $\bm{x}_j$ tend to be similar if $\hat{\bm{h}}_i^\top \hat{\bm{h}}_j$ is close to 1, and dissimilar if $\hat{\bm{h}}_i^\top \hat{\bm{h}}_j$ is close to 0. Considering the completely random case when $\hat{\bm{h}}^\top_i=\hat{\bm{h}}^\top_j=(1/K, ..., 1/K)$ and $\hat{\bm{h}}_i^\top \hat{\bm{h}}_j=1/K$, we choose $1/K$ as a threshold for the effective concordance between $\bm{x}_i$ and $\bm{x}_j$, and define the augmented constraints as $\tilde{\mathcal{S}}=\{(i,j)|\hat{\bm{h}}_i^\top \hat{\bm{h}}_j>1/K\}$ and $\tilde{\mathcal{D}}=\{(i,j)|\hat{\bm{h}}_i^\top \hat{\bm{h}}_j<1/K\}$.  Then we train the metric matrix through
\begin{equation}\label{eqn:metric learn with augmented}
    \begin{split}
        &\min_A \quad Loss(A) \triangleq \frac{1}{|\mathcal{S}|} \sum_{(i,j)\in \mathcal{S}} \|\bm{x}_i-\bm{x}_j\|_A^2 + \frac{1}{ |\tilde{\mathcal{S}}|}\sum_{(i,j)\in \tilde{\mathcal{S}}} w_{ij} \|\bm{x}_i-\bm{x}_j\|_A^2,\\
        &s.t. \quad \frac{1}{|\mathcal{D}|}\sum_{(i,j)\in \mathcal{D}} \|\bm{x}_i-\bm{x}_j\|_A + \frac{1}{ |\tilde{\mathcal{D}}|} \sum_{(i,j)\in \tilde{D}}w_{ij}\|\bm{x}_i-\bm{x}_j\|_A \geq 1, \quad A \succeq 0,
    \end{split}
\end{equation}
where $|\cdot|$ denotes the set cardinality, and 
$w_{ij}=
\frac{K}{K-1}\max\{\hat{\bm h}_i^{\top}\hat{\bm h}_j-\frac{ 1}{ K}, 0\} - K \min\{\hat{\bm h}_i^{\top}\hat{\bm h}_j-\frac{1}{K},0 \}$. Compared with (\ref{eqn:xing}), we involve the augmented similar constraints $\tilde{\mathcal{S}}$ and the dissimilar constraints $\tilde{\mathcal{D}}$ in both $Loss(A)$ and the constraint in (\ref{eqn:metric learn with augmented}),  and we normalize each term by set cardinality to avoid inconsistent scales caused by imbalanced numbers of similar and dissimilar pairs. In addition, we use $w_{ij}\in [0, 1]$ to quantify the certainty of the inference by imposing less weight on the augmented constraints that are similar to random guess, while imposing a large weight on the constraints queried from users and the inferred constraints if their concordance equals 0 or 1. In this way, we are able to fully utilize the information from the total of $n(n-1)/2$ pairs to learn the metric matrix, instead of $|\mathcal{S}|+|\mathcal{D}|$ as in the conventional metric learning methods.

\subsection{Metric aggregation through adaptive penalty}
\label{sub: metric aggregation}

We propose to aggregate the metric matrices learned in each step to extract the underlying significant features by imposing an adaptive penalty on (\ref{eqn:metric learn with augmented}), and capture a clustering-oriented subspace. Note that imposing a penalty on all features simultaneously makes a limited impact on the clustering result, since clustering is invariant to the scale of the elements in the metric matrix. Instead, we impose a selective penalty on a subset of features to increase the relative weights of the significant features over the irrelevant ones.

We denote the minimizer of (\ref{eqn:metric learn with augmented}) at the $t$th step as $A^t, ~t=1,..., T-1$. To select important features, we aggregate the results of the previous $T-1$ steps by imposing a penalty adaptively based on the eigenvalues of $A^t$. In general, the features with smaller eigenvalues on average are less relevant in clustering and thus should have smaller weights.  We let $\bm{r}^t=(r_1,\cdots, r_p),$ be the rank statistics of $p$ eigenvalues from $A^t$ in ascending order, and $\bar{\bm{r}}=\frac{1}{T-1}\sum_{t=1}^{T-1} \bm{r}^t$ be the average rank. To shrink the weights on irrelevant features, we penalize the top $q$ features with the smallest entries in $\bar{\bm{r}}$, {where $q$ is the number of penalized features.} We denote the index set of the penalized features at the $T$th step as $\mathcal{G}^T$, $|\mathcal{G}^T|=q$ . Then for the $T$th step, we train the metric matrix by adding a selective penalty on $A$ through
\begin{equation}\label{eqn:partial penalty full rank}
  \begin{aligned}
        &\hat{A} = \argmin_A \quad Loss(A)+ \gamma \sum_{k\in \mathcal{G}^T}\sigma_k(A),\\
        &s.t. \quad\frac{1}{|\mathcal{D}^T|}\sum_{(i,j)\in \mathcal{D}^T} \|\bm{x}_i-\bm{x}_j\|_A + \frac{1}{ |\tilde{\mathcal{D}}^T|} \sum_{i,j}w_{ij}\|\bm{x}_i-\bm{x}_j\|_A \geq 1, \quad A \succeq 0,
  \end{aligned}
\end{equation}
where $\gamma$ is a tuning parameters and $\sigma_k(A)$ denotes the $k$th eigenvalue of $A$.
The reason we use the rank statistic in determining $\mathcal{G}^T$ instead of using the eigenvalue directly is that the rank statistic is {robust to outliers from the distribution of eigenvalues}, which lowers the risk of incorrectly penalizing significant features. In particular, when the metric matrix is diagonal, the proposed selective penalizing procedure is equivalent to adding an $L_1$ penalty to a subset of the diagonal entries of $A$.
{Different from the nuclear norm penalty which penalizes all eigenvalues of $A$, the selective penalty (\ref{eqn:partial penalty full rank}) only penalizes the eigenvalues in $\mathcal{G}^T$. }
\color{black}

After acquiring $\hat{A}$ through (\ref{eqn:partial penalty full rank}), we solve for the cluster membership by performing pairwise constrained Kmeans (PCKmeans) \cite{basu_active_2004} on the learned linear subspace via
\begin{equation} \label{eqn: metric_PCKmeans}
   \hat{\bm{l}}=\argmin_{\bm{l}} ~~\sum_{i=1}^n \|\bm{x}_i-\bm{\nu}_{\ell_i}\|_{\hat{A}}^2 + \sum_{(i,j)\in \mathcal{S}^T} \mathbbm{1}({\ell_i\neq \ell_j}) + \sum_{(i,j)\in \mathcal{D}^T} \mathbbm{1}({\ell_i= \ell_j}),
 \end{equation} 
where $\bm{\nu}_k = {\sum_{i=1}^n \bm{x}_i \mathbbm{1}(\ell_i=k)}/{\sum_{i=1}^n \mathbbm{1}(\ell_i=k)}$ is the centroid of the $k$th cluster. Here we shrink the search space of the membership $\bm{l}$ by penalizing the cases where the label assignments violate the queried pairwise constraints. Different from the original PCKmeans, we compute the distances between the samples and cluster centers with the learned metric $\hat{A}$ so that irrelevant features are excluded.

\color{black}


\subsection{Active query with minimum expected entropy} 
\label{sub:active_query_selection} 
\label{sub:active_query_with_maximum_expected_cross_entropy}
In this subsection, we introduce an active strategy to select the unlabeled data pairs to query from users.

{We propose to select instances whose neighborhood membership affects the expected posterior distribution of the cluster label assignment $\bm{l}$ most significantly, and to utilize the neighborhood structure \cite{xiong_active_2014} to augment the queries.} This procedure increases the query efficiency by generating more pairwise constraints within a single query due to the transitivity property. Formally, we define a neighborhood as a subset of instances which belong to the same cluster based on the user's feedbacks. Therefore, any pairs within the same neighborhood are similar while any pairs across different neighborhoods are dissimilar. We denote the $m$th neighborhood at the $t$th step as ${N}_m^t$, then for any $\bm{x}_i, \bm{x}_j \in N_m^t$, we have $(i, j)\in \mathcal{S}^t$, and for any $\bm{x}_i \in N_m^t$, $\bm{x}_j \in N_{m'}^t, m\neq m'$, we have $(i,j)\in \mathcal{D}^t$. 

We start with one neighborhood which contains a single instance, and sequentially identify the memberships of the instances outside the existing neighborhoods by querying their similarity with the instances within the neighborhoods. Specifically, we denote the union of the neighborhoods at the $t$th step as $\mathcal{N}^t=N_1\cup\cdots \cup N_{L^t}$, where $L^t\leq K$ is the total number of neighborhoods; then for the next step, we select an $\bm{x}_i \not\in \mathcal{N}^t$ and determine its neighborhood membership  by querying its similarity with one representative from each neighborhood $\bm{x}_j \in N_{j}, j=1,\cdots,N_{L^t}$ sequentially until a similar pair is found. If $\bm{x}_i$ does not belong to any of the neighborhoods, we formulate a new neighborhood as $\{\bm{x}_i\}$ and update $\mathcal{N}$ with $\mathcal{N}^t \cup \{\bm{x}_i\}$ and $L$ with $L^t+1$. Note this query procedure costs at most $K$ queries, but can generate $|\mathcal{N}^t|$ pairwise constraints due to transitivity. In addition, since $|\mathcal{N}^t|$ increases as $t$ grows, we are able to acquire more constraints with the same cost as the query procedure continues.

Next, we introduce a new uncertainty measurement to select unlabeled instance. 
Note that the optimization problem (\ref{eqn: metric_PCKmeans}) can be formulated to maximize the posterior distribution $\rho(\bm{l}|\bm{x})\propto f(\bm{x}|\bm{l})\pi(\bm{l})$, where the likelihood function given the cluster memberships is 
\begin{equation*}
	f(\bm{x}|\bm{l})\propto \exp\left(-\frac{1}{2}\sum_{i=1}^n \|\bm{x}_i-\bm{\nu}_{\ell_i}\|_{{A}}^2\right),
\end{equation*}
and the prior distribution on $\bm{l}$ is
\begin{equation}\label{eqn: exponential prior}
	\pi(\boldsymbol{l}) \propto \exp \left(-\sum_{i, j} V_{(i, j)}\left(\ell_{i}, \ell_{j}\right)\right),
\end{equation}
with
\begin{equation*}
	V_{(i,j)}(\ell_i, \ell_j)=
	\begin{cases}
		\mathbbm{1}(\ell_i\neq \ell_j)\quad &(i,j)\in \mathcal{S},\\
		\mathbbm{1}(\ell_i = \ell_j)\quad &(i,j)\in \mathcal{D},\\
		0 & \text{otherwise}.
	\end{cases}
\end{equation*}

\noindent
We denote the target posterior as $\rho_{*}(\boldsymbol{l}|\boldsymbol{x}) \propto f(\boldsymbol{x}|\boldsymbol{l}) \pi_{*}(\boldsymbol{l})$, where $\pi_*(\bm{l})$ denotes the prior with the labels of all data pairs; and the posterior of the $t$th step as $\rho^{t}(\boldsymbol{l}|\boldsymbol{x}) \propto f(\boldsymbol{x}|\boldsymbol{l}) \pi^{t}(\boldsymbol{l})$, where $\pi^{t}(\boldsymbol{l})$ involves $\mathcal{S}^t$ and $\mathcal{D}^t$ only. Under this framework, the discrepancy between $\rho^{t+1}(\bm{l}|\bm{x})$ and $\rhoo$ relies on $\pi^{t+1}(\bm{l})$, which is determined by the query strategy.

We propose to select $\bm{x}_i \not\in \mathcal{N}^t$ whose neighborhood membership is expected to make the posterior distribution closest to the underlying truth via minimizing the Kullback–Leibler divergence (KL-divergence): 
\begin{equation}\label{eqn: KL0}
	{\bm{x}}_i^* 
	=\argmin_{\bm{x}_i \not\in \mathcal{N}^t}\sum_{\bm{l}\in\Omega}\rhoo \log\left(\frac{\rhoo}{ \rho^+_i(\bm{l}|\bm{x})}\right)=\argmin_{\bm{x}_i \not\in \mathcal{N}^t} -\sum_{\bm{l}\in\Omega}\rhoo \log \rho^+_i(\bm{l}|\bm{x}),
\end{equation}
where $\rho^+_i(\bm{l}|\bm{x})$ denotes the posterior distribution after determining the neighborhood membership of $\bm{x}_i$.
However, since both $\rho_0$ and the true membership of $\bm{x}_i$ are unobserved, we cannot solve (\ref{eqn: KL0}) directly. 
Instead, we consider the following approximation:
\begin{equation} \label{eqn: expected entropy}
	{\bm{x}}_i^* = \argmin_{\bm{x}_i \not\in \mathcal{N}^t} -\sum_{m=1}^{L^t} P^t(\ell_i=m) \sum_{\bm{l}\in\Omega}\rho^+_{im}(\bm{l}|\bm{x}) \log \rho^+_{im}(\bm{l}|\bm{x}),
\end{equation}
where $\rho^+_{im}(\bm{l}|\bm{x})$ denotes the posterior distribution assuming $\bm{x}_i\in N_m$. The active query strategy (\ref{eqn: expected entropy}) can be interpreted as a minimization of the expected entropy of the posterior distribution when new constraints are added, which is equivalent to selecting the instance whose neighborhood membership is the most uncertain based on the information at the $t$th step. The neighborhood structure is shown to be effective in the normalized point-based uncertainty (NPU) \cite{xiong_active_2014}. However, the NPU considers the uncertainty decrease only based on the queried instance, while the proposed method (\ref{eqn: expected entropy}) measures the uncertainty decrease over the entire dataset. Therefore, the proposed criterion estimates the information gain from the new query holistically and globally, and thus is able to select the pivotal instances.

\section{Algorithm and implementation} 
\label{sec:implementation}

In this section, we introduce the algorithms to solve the query augmentation problem in (\ref{eqn:impute_H}), the metric learning with selective penalty in (\ref{eqn:partial penalty full rank}), and the active query strategy in (\ref{eqn: expected entropy}).

We adopt the alternating direction method of multipliers (ADMM) method to solve (\ref{eqn:impute_H}), and decompose the optimization problem (\ref{eqn:impute_H}) into several subproblems that can be solved more easily. {We leave the details in Supplementary Materials because of space limit.}

For the selective penalty optimization (\ref{eqn:partial penalty full rank}), we first consider the diagonal case. Denote the diagonal entries of $A$ as $\bm a$, then the selective penalty is equivalent to penalizing a subgroup of entries in $\bm{a}$ directly. We consider the equivalent form of (\ref{eqn:partial penalty full rank}) such that the constraint is linear regarding $\bm{a}$:
\begin{equation}
  \begin{split}
  \label{eqn: partial penalty diagonal case}
  &\hat{\bm{a}}^T=\arg\max_{\bm{a}} \frac{1}{|\mathcal{D}^T|}\sum_{(i,j)\in \mathcal{D}^T} \sqrt{\sum_{m=1}^p a_m(x_{im}-x_{jm})^2} + \frac{1}{ |\tilde{\mathcal{D}}^T|} \sum_{i,j}w_{ij}\sqrt{\sum_{m=1}^p a_m(x_{im}-x_{jm})^2} \\ 
  &s.t. \quad \frac{1}{|\mathcal{S}^T|} \sum_{(i,j)\in S^T} \sum_{m=1}^p a_m(x_{im}-x_{jm})^2 + \frac{1}{ |\tilde{\mathcal{S}}^T|}\sum_{i,j}\sum_{m=1}^p w_{ij} a_m(x_{im}-x_{jm})^2  + \gamma \sum_{p\in \mathcal{G}^T} a_p  \leq 1, \quad a_p \geq 0.
  \end{split}
\end{equation}
The above function can be maximized using the projected gradient descent method.

For the general case, we separate the optimization procedure into two steps. We first seek the best subspace in which data can be clustered more efficiently, and transform the features into this linear subspace accordingly.  We then train the weight of the transformed features similar as the diagonal case. Mathematically, consider the decomposition $A=PDP^{\top}$ where $P\in \mathbb{R}^{p\times p}$ is an orthonormal matrix and $D$ is a diagonal matrix. We first let $\hat{P}$ be the eigenvectors of the solution to $(\ref{eqn:partial penalty full rank})$ with $\gamma=0$, which  can be completed using projected gradient descent. Then we transform each data point $\bm{x}_i$ by $\hat{P}^{\top}\bm{x}_i$ and the problem degenerates to the diagonal case $(\ref{eqn: partial penalty diagonal case})$. We denote the solution with selective penalty as the diagonal weighting matrix $\hat{D}$, and assemble the target metric matrix as $\hat{A}=\hat{P}\hat{D}\hat{P}^{\top}$.
\color{black}

Next, we implement the active query strategy (\ref{eqn: expected entropy}) with the neighborhood structure. Notice that computation of the exact expected entropy in (\ref{eqn: expected entropy}) requires the enumeration of all possible membership assignments over $\Omega$, which is computationally infeasible. 
Alternatively, we propose to approximate the expected entropy by taking the summation of the expected entropy for each unlabeled pairs, which contains at most $|\mathcal{U}|\leq n(n-1)/2$ terms. Furthermore, 
instead of considering the posterior distribution of each pair directly, we estimate the posterior distribution based on the neighborhood membership of each data point to simplify computation.

Specifically, we let $R^t\in \mathbb{R}^{n\times L^t}$ be the neighborhood membership matrix, where $r_{im}^t=P(\bm{x}_i\in N_m^t)$. Given that the data is sampled independently, the probability that $\bm{x}_i$ and $\bm{x}_j$ belong to the same neighborhood can be computed as $p_{ij}^t=\sum_{m=1}^{L^t}P(\bm{x}_i\in {N}^t_m, \bm{x}_j\in {N}^t_m)=\sum_{m=1}^{L^t}r^t_{im}r^t_{jm}$. Then we approximate the entropy in (\ref{eqn: expected entropy}) by
\begin{equation*}
  Q(R^t)=-\sum_{(i,j)\in \mathcal{U}} \left\{p^t_{ij}\log p^t_{ij} + (1- p^t_{ij})\log (1-p^t_{ij}) \right\},  
\end{equation*}
where $R^t$ is implicitly included in each $p^t_{ij}$ and is omitted in the expression for notation simplicity. The expected entropy by identifying the neighborhood membership of $\bm{x}_i$ is then
\begin{equation}\label{eqn: uncertainty}
  u^t(\bm{x}_i)=\sum_{m=1}^{L^t} r^t_{im}Q(\tilde{R}^{t+1}_{-i,m}),
\end{equation}
where $\tilde{R}^{t+1}_{-i,m}\in \mathbb{R}^{n\times L^{t+1}}$ is defined elementwise by
$$
  \tilde{r}^{t+1}_{ij}=
  \left\{
  \begin{array}{ll}
  r_{kj}^t, &\text{ if } k\neq i, \\
  0,       &\text{ if } k=i \text{ and } j \neq m, \\
  1,       &\text{ if } k=i \text{ and } j = m. 
  \end{array}
  \right.
$$
That is, $\tilde{R}^{t+1}_{-i,m}$ denotes the neighborhood membership matrix assuming that $\bm{x}_i$ belongs to the $m$th neighborhood, and $u^t(\bm{x}_i)$ estimates the expected entropy after obtaining the neighborhood membership of $\bm{x}_i$. We then select 
$
  {\bm{x}}^{*} = \argmin_{\bm{x}_i\notin \mathcal{N}^t} u^t(\bm{x}_i).
$

We estimate $r^t_{im}$ by a random forest trained on the clustering result with the learned metric matrix at the $t$th step. The random forest is computationally efficient without assuming the explicit form of $f(\bm{x}|\bm{l})$, making the model more flexible for general cluster structures. The random forest has been applied successfully in previous studies \cite{xiong_active_2014,shi2006unsupervised} for unsupervised clustering tasks, especially for quantifying the uncertainty of memberships. The complete algorithm is summarized in Algorithm \ref{alg:metric_aggregation}.
\color{black}

		\begin{algorithm}[htp]\label{algorithm: entire algorithm}
		  \caption{Query-augmented active clustering with metric aggregation}
		  \label{alg:metric_aggregation}
		  {Input}: Data $\{\bm{x}_i\}$, {budget $B$}, number of clusters $K$. \\
		  {Output}: Cluster label $\bm{l}$. \\
		  {Initialization}: Single neighborhood $\mathcal{N}=\{N_1\}, N_1=\{\bm{x}_1\}$, where $\bm{x}_1$ is randomly selected. Let $\mathcal{G}^0=\mathcal{S}=\mathcal{D}=\varnothing$, $A^0=\bm{I}_p$ and {$t=0$}.
		  \begin{enumerate}[leftmargin=*,label={\arabic*.}]
		  
		  	\item (\textit{Active metric learning}){{While} $ \sum_t^T b^t \leq B$}, {repeat}:
			  \begin{enumerate}[label={(\alph*)}]
			    \item (\textit{Active query}) Train a random forest to estimate the neighborhood membership matrix $R^t$. Select the most informative instance $\bm{x}^*$ to minimize (\ref{eqn: uncertainty}). Sort $\{N_i^t\}'s$ in descending order of $p(\bm{x}^*\in N_i^t)$. {Let $b^t=0$.}
			    \item {Query $\bm{x}^*$ against an instance $x_i\in N_i^t$, update $\mathcal{S}$ or $\mathcal{D}$ according to the feedback, {$b^t\leftarrow b^t+1$}.}
			    \item 
			    {Repeat step (b) for the rest of the neighborhoods until a similar link between $\bm{x}^*$ and $\bm{x}_i$ is provided by user or {$\sum_t^T b^t = B$}. Let $N_i^t\leftarrow N_i^t\cup \{\bm{x}^*\}$. If no similar link is found, treat $\bm{x}^*$ as a new neighborhood. Let $N^*\leftarrow \{\bm{x}^*\}$ and $\mathcal{N}^t\leftarrow \mathcal{N}^t \cup \{N^*\}$.}

			    \item (\textit{Metric update}) Augment the pairwise constraints by solving (\ref{eqn:impute_H}) using the ADMM algorithm . Train metric $A^t$ with the augmented queries by (\ref{eqn:metric learn with augmented}). {Let $t\leftarrow t+1$.}
			    
			  \end{enumerate}

		  \item  (\textit{Metric aggregation}) Compute $\mathcal{G}^T$ based on $\{A^t\}_{t=1}^T$, solve for $\hat{A}$ with selective penalty (\ref{eqn:partial penalty full rank}).
		  \item (\textit{Semi-supervised clustering}) Cluster the instances with PCKmeans (\ref{eqn: metric_PCKmeans}) based on the learned metric $\hat{A}$ and the acquired pairwise constraints $\mathcal{S}$ and $\mathcal{D}$.
		  \end{enumerate}
		\end{algorithm}

	The total computational complexity of Algorithm \ref{alg:metric_aggregation} is $\mathcal{O}\{T(n_{\tau} n \log n+n^{2} K+ \frac{1}{\epsilon^{2}} n K^{3} + $$\frac{1}{\epsilon}p^3) + npK t_{pck}\}$, where $n_\tau$ is the number of trees in the random forest used during active query; $\epsilon$ is the predetermined error bound for constraint augmentation and metric learning; and $t_{pck}$ is the iteration number of PCKmeans.	Specifically, for a single query step, the active query selection costs $\mathcal{O}\left(n_{\tau} n \log n+n^{2} K\right)$, where $n_{\tau} n \log n $ is the complexity of training a random forest \cite{breiman2003rf} and $n^{2} K$ is the complexity to compute the information criterion $u^t(\bm{x}_i)$. In addition, the query augmentation costs $\mathcal{O}\left(\frac{1}{\epsilon^{2}} n K^{3}\right)$, and the metric learning procedure requires $\mathcal{O}(\frac{1}{\epsilon}p^2)$ or $\mathcal{O}(\frac{1}{\epsilon}p^3)$ for the diagonal and the non-diagonal $A$, respectively. Finally, the PCKmeans costs $np Kt_{pck}$. Empirically, training a random forest with 50 trees takes 0.05 seconds and one loop in the simulation setting with $p=35, n=300$ and $K=5$ costs 8 seconds on an Intel 4-Core i7-8650U CPU at 1.90GHz.


	In the following, we provide a brief discussion on the selection of tuning parameters in Algorithm \ref{alg:metric_aggregation}, i.e., $\lambda$ in constraint augmentation (\ref{eqn:impute_H}), $\gamma$ and $q$ in the selective penalty (\ref{eqn:partial penalty full rank}). In practice, $\lambda$ is selected by a 5-fold cross validation based on the labeled pairwise constraints with a grid search on $[0,1]$ after each loop of step (1) in Algorithm \ref{alg:metric_aggregation}. On the other hand, we select $\gamma$ by maximizing the Calinski-Harabasz index \cite{calinski1974dendrite}, which evaluates the clustering results by the ratio of the between-cluster variance and the within-cluster variance obtained from the PCKmeans. Different from $\lambda$, we tune $\gamma$ only in Step 2 of Algorithm \ref{alg:metric_aggregation}. Finally, $q$ can be selected based on the unpenalized metric learning from Step 1 in Algorithm \ref{alg:metric_aggregation}. Specifically, $q$ is selected by the elbow point corresponding to the average eigenvalues of the metric matrices $\{A^t\}_{t=1}^T$, which are trained without penalty. We refer the readers to the Supplementary Materials for more details on the parameter selection in the simulation and real data experiments.
\color{black} 

\section{Theory} 
\label{sec:theory}
In this section, we introduce theoretical results for the proposed active clustering method. We first show the advantage of incorporating the augmented pairwise constraints in the metric learning step, and next demonstrate the improvement of the active query strategy over the passive learning approach.

We formulate the metric learning into the semi-supervised learning framework which consists of $(\bm{x}_i, \bm{x}_j)$ as pairs of data, and $y_{ij}$'s as labels. Our goal is to learn a binary classifier parametrized by the metric matrix $A$ trained by the pairwise constraints as labeled data.  The number of queries required to achieve a certain prediction accuracy without considering the unlabeled data can be derived by the {VC dimension \cite{devroye2013probabilistic}. The VC dimension of a function space $\mathcal{C}$ is the maximum number of arbitrarily labeled points that can be classified correctly by the functions in $\mathcal{C}$}. However, utilizing the underlying clustering data structure provides additional implicit constraints and reduces the searching space of the target classifier, which requires fewer queries and therefore accelerates the training process. This is achieved by imposing a penalty on the incompatibility of unlabeled pairs with the metric through the augmented labels $\tilde{\mathcal{S}}$ and $\tilde{\mathcal{D}}$ in (\ref{eqn:metric learn with augmented}). The proposed method is able to minimize both the classification error and the incompatibility simultaneously.

Specifically, the loss function in (\ref{eqn:xing}) can be written as 
\begin{equation*}
\frac{1}{|\mathcal{S}|+|\mathcal{D}|}\sum_{i,j\in \mathcal{S}\cup \mathcal{D}}\mathbbm{1}\{y_{ij}=0\}(1-\|\bm{x}_i-\bm{x}_j\|_A) + \mathbbm{1}\{y_{ij}=1\}(\|\bm{x}_i-\bm{x}_j\|_A^2-1),	
\end{equation*}
and is a surrogate function to 
\begin{equation*}	
\hat{e}(h_A)=\frac{1}{|\mathcal{S}|+|\mathcal{D}|}\sum_{i,j\in \mathcal{S}\cup \mathcal{D}}(2y_{ij}-1) h_A(\bm{x}_i, \bm{x}_j), 
\end{equation*}
where $h_A(\bm{x}_i, \bm{x}_j)=\text{sign}(\{\|\bm{x}_i-\bm{x}_j\|_A^2-1\})$. We denote the joint distribution of $(\bm{x}_i, \bm{x}_j)$ as $F$, then $\hat{e}(h_A)$ is the empirical estimator of the error rate for the labeled data $e(h_A)=P_{(\bm{x}_i, \bm{x}_j)\sim F}(h_A(\bm{x}_i, \bm{x}_j)\neq 2y_{ij}-1)$. {In addition, we define the \textit{incompatibility} between $h_A$ and $F$ as }
\begin{equation}\label{eqn: incompatibility}
e_u(h_A)=E_{(\bm{x}_i, \bm{x}_j)\sim F}~\chi(h_A, \bm{x}_i, \bm{x}_j), 	
\end{equation}
where  $\chi(h_A, \bm{x}_i, \bm{x}_j) = P(\ell_i=\ell_j)\|\bm{x}_i-\bm{x}_j\|_A^2- P(\ell_i\neq \ell_j)\|\bm{x}_i-\bm{x}_j\|_A$. Intuitively, $e_u$ measures the average proximity among data pairs weighted by the probability of being from the same cluster under the metric $A$, and $e_u$ is small if the metric captures the important features. 
The empirical estimator of $e_u$ is 
\begin{equation*}
\hat{e}_u(h_A)=\frac{2}{n(n-1)}\sum_{i,j} w_{ij}\mathbbm{1}\{\hat{\bm h}_i^{\top}\hat{\bm h}_j-{ 1}/{K}>0 \} \|\bm{x}_i-\bm{x}_j\|_A^2 - w_{ij}\mathbbm{1}\{\hat{\bm h}_i^{\top}\hat{\bm h}_j- { 1}/{K}<0 \} \|\bm{x}_i-\bm{x}_j\|_A	,
\end{equation*}
where $w_{ij}$ and $\hat{\bm{h}}_i$ are defined as in (\ref{eqn:partial penalty full rank}). Then the proposed augmented metric learning (\ref{eqn:partial penalty full rank}) is equivalent to minimizing both $\hat{e}(h_A)$ and $\hat{e}_u(h_A)$ at the same time.


Furthermore, {we denote the function space of $h_A$ as $\mathcal{C}=\{h_A: A\in\mathbb{R}^{p\times p}\text{ and } A \text{ is semidefinite}\}$}. To quantify the complexity of $\mathcal{C}$ regarding the binary classification task for data sampled from $F$, {we can draw a sample of size $n$ independently from $F$ and classify it with the functions in $\mathcal{C}$. The expected number of label assignments that can be correctly classified is denoted as $S^{\mathcal{C}}_{F}(n)$.}
Note that $S^{\mathcal{C}}_{F}(n)$ is a distribution-dependent complexity measure of $\mathcal{C}$. In general, a larger $S^{\mathcal{C}}_{F}(n)$ implies a larger function space $\mathcal{C}$. In addition, we let $\mathcal{C}_{\chi}(t)=\{h_A\in \mathcal{C}: e_u(h_A)\leq t\}$ be the function space whose {{incompatibility with $F$ as defined in (\ref{eqn: incompatibility})}} is bounded by $t$, where $t$ is a positive constant.

In the following theorem, we show the classification accuracy achieved by the proposed query-augmented metric learning method given the increasing number of pairwise constraints.

\begin{theorem}\label{theorem: rate}
	Given any $\epsilon, s, t> 0$, and the number of pairwise constraints $n_l=|\mathcal{S}|+|\mathcal{D}|$, assume that
	\begin{equation*}
	  \frac{n(n-1)}{2}-n_l = \mathcal{O}\left(\frac{p+1}{\epsilon^2}\log \frac{1}{\epsilon} +\frac{1}{\epsilon^2}\log \frac{2}{\delta}\right),
	\end{equation*}
	where $p$ is the dimension of $A$ and
	\begin{equation*}
		\delta = 8 S^{\mathcal{C}_{\chi}(t+2 \epsilon)}_{F}(2n_l)\exp\left(-\frac{1}{2} \epsilon n_l \right).
	\end{equation*}
	Then for all $h_A \in \mathcal{C}$ with $\hat{e}(h_A)\leq s$ and $\hat{e}_u(h_A)\leq t+\epsilon$, we have $P(e(h_A) \leq s + \epsilon) \geq 1-\delta$.
\end{theorem}
\noindent
Since $\mathcal{C}_{ \chi}(t+2\epsilon)$ is a subset of $\mathcal{C}$, we have
\begin{equation}\label{eqn: growth function bound}
	S^{\mathcal{C}_{\chi}(t+2 \epsilon)}_{F}(2n_l)\leq S^{\mathcal{C}_{\chi}}_{F}(2n_l)\leq \left(\frac{2en_l}{p+1}\right)^{p+1},
\end{equation}
if $2n_l > p+1$. {Therefore, $\delta$ converges to 0 as $n_l$ increases to infinity, indicating that the classification accuracy using the learned metric converges to the optimal accuracy of $h_A\in \mathcal{C}$ with a probability approaching 1.
The second inequality in (\ref{eqn: growth function bound}) is derived from the relation between the growth function and the VC dimension \cite{vapnik2013nature}, where the growth function is the supremum of $S_F^\mathcal{C}(n)$ among all $F$'s and the VC dimension equals $p+1$ in our case. }
In addition, note that by using the labeled data only, the convergence rate is
\begin{equation*}
	\delta_0=4 S^{\mathcal{C}}_{F}(2n_l)\exp\left(-\frac{1}{2} \epsilon n_l \right).
\end{equation*}
Thus, we have $\delta < \delta_0$ if $S_{F}^{\mathcal{C}_{\chi}(t+2 \epsilon)}\left(2 n_{l}\right)< \frac{1}{2} S_{F}^{\mathcal{C}}\left(2 n_{l}\right)$, which can be satisfied if the label of at least one pair from $\mathcal{U}$ can be augmented correctly. With this additional condition, we are able to achieve a faster convergence rate by incorporating the unlabeled pairs with the augmented information.

Next, we show an improvement on utilizing the proposed active query strategy compared with random selection.  In the following, we denote the posteriors of the membership assignment from the random query and active query after acquiring the membership of one extra instance at the $t$th step as $\rho_{random}$ and $\rho_{active}$, respectively, and denote the underlying distribution $\rho_0$ as in Section \ref{sub:active_query_with_maximum_expected_cross_entropy}. We denote the size of the clusters as $\bm{\alpha}=(\alpha_1,\cdots,\alpha_K)$, and the size of the neighborhoods at the $t$th step as $\bm{\beta}^t=(\beta_1^t, \cdots, \beta_{L^t}^t)$. Note that $\bm{\alpha}$ does not change with $t$ and $\sum_i^K \alpha_i=n$.
We formulate the constrained clustering process from a Bayesian perspective in that the constraints are added as a prior in the form of (\ref{eqn: exponential prior}). The following conditions are assumed to hold:
\begin{enumerate}
	\item [(I)] The likelihood of $\bm{x}_i$'s follows $P(x_i|\ell_i=k)\propto$$ \exp \left\{-\left(\frac{\|x_{i}-\mu_{k}\|_A}{\sigma}\right)^{d}\right\}$, where $d$ is a positive constant, $\mu_k$ is the center of the $k$th cluster and $\sigma^2=\mathcal{O}\{Var(X_i^2|y_i=k)\}$.
	\item [(II)]  The distances between any two cluster centers are equal, which is denoted as $r=\|\mu_{i}-\mu_{j}\|_A$.
	\item [(III)] There is one and only one neighborhood in each cluster and thus $L^t=K$.
\end{enumerate}
Condition (I) specifies a unimodal cluster structure, which is a common assumption in the analyses of probabilistic clustering models \cite{marlin2012unsupervised,rodrigues2014probabilistic,lu2007penalize}. Note that the Gaussian distribution is a special case when $d=2$. Conditions (II) and (III) are assumed only for simplicity of notations in proofs. In fact, we can generalize our conclusions without (II) by letting $r$ be the shortest distance between any two cluster centers, and the results and proofs remain the same. Condition (III) facilitates a convenient formulation of the posterior probability in the following theorem. A more general case can be proved similarly without (III), but requires more on combinatorial analysis, which might be unnecessary in illustrating the magnitude of the tail probability. 

\begin{theorem}\label{theorem:KL}
Given the {neighborhoods} at the $t$th step $\mathcal{N}^t$ and $n>|\mathcal{N}^t|$, assume conditions (I)-(III) hold, then we have
\begin{equation*}
	{KL(\rho_0||\rho_{active})
		\leq KL(\rho_0||\rho_{random})},
\end{equation*}
with probability at least $1-\epsilon$, where
\begin{equation}\label{eqn: convergene rate}
	\begin{split}
	\epsilon =  \mathcal{O}_p\left[ n\log K \exp\left\{-n\left(\frac{r}{\sigma}\right)^d\right\}\left\{\xi(\bm{\beta}^t) - \xi(\bm{\alpha}) \right\}\right],
	\end{split}
\end{equation}
with
\begin{equation*}
\begin{gathered}
	\xi(\bm{x}) =  \left\{\sum_{\substack{i,j\\i\neq j}}^K x_i \exp\left(-x_i-x_j\right)\right\}^n.\\
\end{gathered}
\end{equation*}
\end{theorem}
Theorem \ref{theorem:KL} implies that the discrepancy between the underlying true distribution and the updated posterior distribution is smaller based on the active query strategy than random selection with a probability close to 1. 
The tail probability $\epsilon$ in (\ref{eqn: convergene rate}) depends on the sample size $n$, neighborhood size $\bm{\beta}^t$ and the underlying cluster structure. 
In particular, $\epsilon$ converges to 0 as $n$ increases if $\sum_{\substack{i,j\\i\neq j}}^K \beta_i^t \exp\left(-\beta_i^t-\beta_j^t\right)<1$, {which can be satisfied when the sample size in each neighborhood is large enough such that $ \exp(2 \min_i \beta_i^t)/(\min_i \beta_i^t)\geq \frac{K(K-1)}{2}$.} Furthermore,  $\epsilon$ decreases exponentially with $n$ in a convergence rate bounded by $\min_i \beta_i^t$ and $\max_i \beta_i^t$, since $K^{2n}(\max_i \beta_i^t)^n \exp(-2n \max_i \beta_i^t) \leq \xi{(\bm{\beta}^t)} \leq K^{2n}(\min_i \beta_i^t)^n \exp(-2n \min_i \beta_i^t)$. 
When $n$ is fixed, $\epsilon$ can still converge to 0 if we add pairwise constraints and decrease the value of $\xi(\bm{\beta}^t) - \xi(\bm{\alpha})$. Notice that $\beta_i<\alpha_i$ for any $i=1,\cdots,K$, {therefore we have $\xi(\bm{\beta}^t) - \xi(\bm{\alpha})\geq 0$, and $\xi(\bm{\beta}^t) - \xi(\bm{\alpha})=0$ when all pairs are queried.} In addition, $\epsilon$ decreases if the clusters are more separable (i.e., $({r}/{\sigma})^d$ is larger). Here $({r}/{\sigma})^d$ can be interpreted as the signal-noise ratio in the clustering task, where $r$ measures the closeness among clusters, while $\sigma$ and $d$ measure the density of data points around the cluster center.

To better illustrate the magnitude of the tail probability, we consider the balanced cluster and balanced neighborhood case:

\begin{corollary}\label{corollary: KL0}
Assume the conditions for Theorem \ref{theorem:KL} hold and $\beta_1=\cdots=\beta_K=\beta$, $\alpha_1=\cdots=\alpha_K=\alpha$, then
\begin{equation*}
	{KL(\rho_0||\rho_{active})
		\leq KL(\rho_0||\rho_{random})},
\end{equation*}
with probability at least $1-\epsilon$, where
\begin{equation}\label{eqn: epsilon balanced}
	\begin{split}
	\epsilon =  \mathcal{O}_p\left\{ n K^{2n}\log K \exp\left[-n\left(\frac{r}{\sigma}\right)^d\right]\left[\beta^n \exp(-2\beta n) - \alpha^n \exp(-2\alpha n) \right]\right\}.
	\end{split}
\end{equation}
\end{corollary}
The probability (\ref{eqn: epsilon balanced}) shows that $\epsilon$ converges to 0 exponentially as $n$ grows if $\exp(2\beta)/\beta > K^2$, which indicates that more pairwise constraints are needed to ensure convergence when there are more clusters. In addition, the convergence of $\epsilon$ is faster when $\beta$ is larger. That is, as the number of queried constraints increases, we are more confident that the proposed active query strategy selects more informative pairs than a passive strategy would.

Although there are some explorations of active supervised learning \cite{balcan2009agnostic} and active hierarchical clustering \cite{eriksson2011active}, most existing active clustering methods with probabilistic models do not provide theoretical properties to the best of our knowledge. In this paper, we introduce a new framework to measure the improvement of utilizing the proposed active strategy over passive selection in terms of the KL divergence. The conclusion of Theorem \ref{theorem:KL} presents theoretical guidance on when the active strategy is effective, and illustrates how model structures such as the number of clusters and separability of clusters affect the convergence rate. Another contribution of Theorem \ref{theorem:KL} is to incorporate the neighborhood information in the analysis, which was proved to be efficient algorithmically but was never investigated theoretically before. Our theoretical properties further justify the rationale behind the proposed query strategy. In addition, we provide a unified criterion on learning efficiency, which can be generalized to evaluate the efficiency of other active clustering strategies.

Theorem \ref{theorem:KL} compares the proposed method with the random query in one single step. In the following theorem, we compare the proposed method with a non-random selection strategy after $T$ queries. 

We define a non-random selection strategy as follows. In Step 1, we perform the K-means clustering. In Step 2, we select the top $T$ most uncertain pairs to query based on the clustering result in Step 1. That is, we select the $T$ pairs with similarity probability closest to $0.5$. The above query method does not involve a sequential query procedure and can be completed in two steps. Therefore, we refer to this simple non-random selection method as the \textit{two-step} strategy, and denote the posterior distribution of the cluster label at the $t$-th step using the two-step strategy as $\rho_{ts}^t(\bm{\ell}|\bm{x})$, accordingly. The comparison between the two-step strategy and the proposed strategy is established in Theorem \ref{theorem:two-step}.

\begin{theorem}\label{theorem:two-step}
	Under the generalized Gaussian distribution Condition (I), for $K^2 < T < Kn$, we have
	\begin{equation}\label{eqn:KL_discrepancy}
		E\left\{KL(\rho_*||\rho_{active}^T) - KL(\rho_*||\rho_{ts}^T)\right \}
		\leq 
		C_1 \sqrt{T} - C_2 T(1-e^{-\frac{2T}{K^2}}),
	\end{equation}
	where $C_1=K\sigma /\sqrt{\underbar{$r$}}$, $C_2 = e^{-(\underbar{$r$}/\sigma)^d}({K-1})/{K}$, and
	$\underbar{$r$} = \min_{i,j} \|\bm{\mu}_i-\bm{\mu}_j\|_A$.
\end{theorem}
Here we require that $T<Kn$ to ensure that not all the pairs have been labeled in the proposed method. Otherwise, $\rho^t_{active}=\rho_*$ and $E\left\{KL(\rho_*||\rho_{active}^T) - KL(\rho_*||\rho_{ts}^T)\right \}$ $= - $ $ E\left\{KL(\rho_*||\rho_{ts}^T)\right \} < 0$.

Theorem \ref{theorem:two-step} compares the Kullback-Leibler divergence of the two methods after $T$ queries, cumulatively. The result of Theorem \ref{theorem:two-step} implies that, with a sufficiently large budget $T>K^2$, we have $E\{KL(\rho_*||\rho_{active}^T)\} < E\{KL(\rho_*||\rho_{ts}^T)\}$; i.e., the posterior distribution of the label $\bm{l}$ using the proposed active query strategy is expected to be closer to $\rho_*$ than the two-step method. Moreover, the upper bound in (\ref{eqn:KL_discrepancy}) indicates that the discrepancy between the proposed method and the two-step method increases as $T$ grows. 
Meanwhile, this discrepancy also depends on the variance $\sigma^2$, the cluster number $K$ and the minimum center distance $\underbar{$r$}$. The result in (\ref{eqn:KL_discrepancy}) demonstrates that the gain of the proposed method is more significant with a larger signal-noise ratio $\underbar{$r$}/ \sigma$ and a smaller $K$.

We remark that the result in (\ref{eqn:KL_discrepancy}) can also be interpreted as the discrepancy of the prior information gain from the labeled data pairs between the two methods: the cumulative information of the two-step grows in $\mathcal{O}(\sqrt{T})$, while the cumulative information of the proposed method grows in $\mathcal{O}(T)$. This is because the two-step method evaluates the informativeness of pairs based on the initial clustering result only, which might not be accurate or effective for the subsequent query steps. Consequently, the information gain of each selected pair may decrease quickly as querying proceeds. In contrast, the proposed method updates the information criterion at each step, and selects the pairs which benefit the posterior distribution at the current step. This sequential updating strategy mitigates the loss in information gain of queried pairs in the subsequent steps, which also facilitates the accumulation of prior information to increase in a linear order. Moreover, the proposed method incorporates more labeled pairs than the two-step method by utilizing the neighborhood structure and the transitivity property, which further enhances cumulative information gain.

The proofs of the theoretical results in this section are provided in the Supplementary Materials.

\color{black}
\section{Simulations} 
\label{sec:simulations}
In this section, we illustrate the advantages of the proposed metric learning method and the active query strategy through simulation settings.
\subsection{The advantages of incorporating augmented constraints} 
\label{sub:the_effectiveness_of_incorporating_}
\begin{figure}[htbp]
  \begin{minipage}[t]{0.48\textwidth}
    \includegraphics[width=\linewidth]{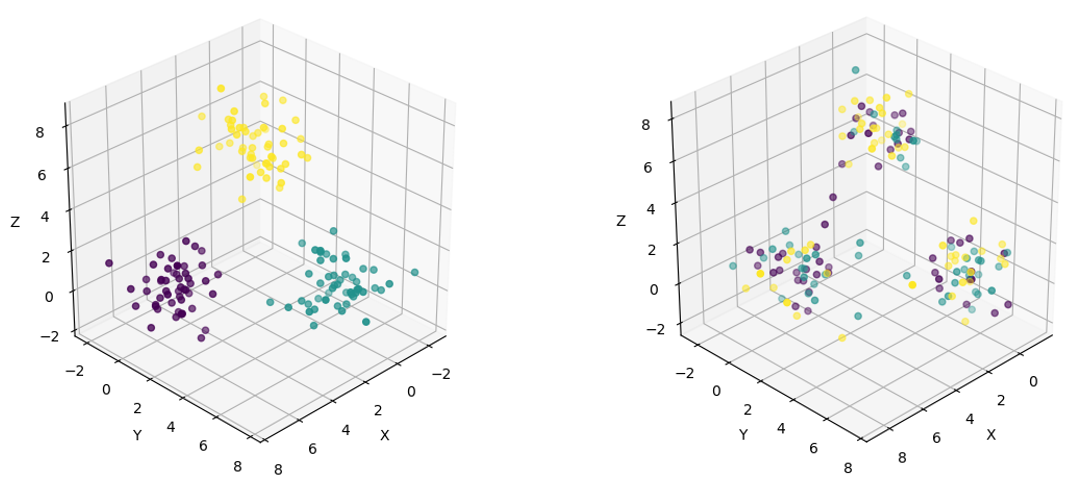}
    \caption{Illustration of the simulated data with $K=p_1=p_2=3$, showing the first three dimensions (\textit{left}) and the last three dimensions (\textit{right}). The cluster membership is determined by the first three dimensions, illustrated by different colors.}
    \label{fig:illustration of simulation}
  \end{minipage}
  \hspace{7mm}
  \begin{minipage}[t]{0.47\textwidth}
    \includegraphics[width=\linewidth]{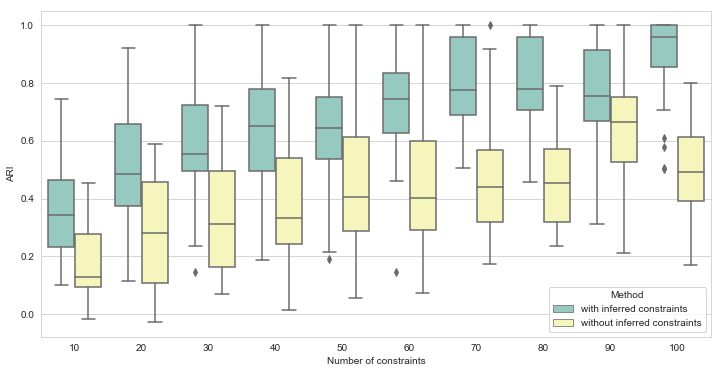}
    \caption{The ARI comparisons of the simulation setting with $p_1=6, p_2=3$ and $c=5$ using random queries with (\textit{green}), and without (\textit{yellow}) augmented constraints, based on 30 replications for each number of constraints.}
    \label{fig:compare H_hat}
  \end{minipage}
\end{figure}

We first demonstrate the advantage of incorporating the augmented constraints (\ref{eqn:metric learn with augmented}). The data points are generated as $\bm{x}^\top=\left\{(\bm{x}^{(1)})^\top, (\bm{x}^{(2)})^\top\right\}$, where $\bm{x}^{(1)} \in \mathbb{R}^{p_1}$ includes the significant features which determine the cluster memberships, and $\bm{x}^{(2)} \in \mathbb{R}^{p_2}$ include irrelevant features. 
Specifically, $\bm{x}^{(i)}$ is sampled from a Gaussian mixture model, i.e., $\bm{x}^{(i)}|z^{(i)}\sim \mathcal{N}(\bm{\mu}^{(i)}_{z^{(i)}}, \bm{I}_{p_i})$, for $i=1, 2$, where $z^{(i)}$ is uniformly sampled from $\{1,\cdots,p_i\}$, and $\bm{\mu}^{(i)}_{z^{(i)}}=c\left(\mathbbm{1}\left\{z^{(i)}=1\right\}, \mathbbm{1}\left\{z^{(i)}=2\right\}, \cdots, \mathbbm{1}\left\{z^{(i)}=p_i\right\}\right)$; i.e., all elements are zero except that the ${z^{(i)}}$th element equals $c$. Here $c$ denotes the distance between the center of the clusters and the origin. A larger $c$ implies a clearer cluster structure.
\color{black}
We let the cluster label $\ell=z^{(1)}$ and the number of clusters $K=p_1$, so the cluster memberships are fully determined by the first $p_1$ features. An illustration of the simulation data with $K=p_1=p_2=3$ is shown in Figure \ref{fig:illustration of simulation}.

In this experiment, we select $p_1=6$, $p_2=3$ , $c=5, n=60$ and $K=6$. We train the metric matrix $A$ with randomly selected pairwise constraints and compare the clustering performance with or without augmented constraints $\tilde{\calS}$ and $\tilde{\calD}$ from (\ref{eqn:metric learn with augmented}), which is evaluated by the adjusted random index (ARI) \cite{rand1971objective}. A higher ARI indicates a clustering result more consistent with the true cluster memberships. Figure \ref{fig:compare H_hat} shows the boxplots of ARI with different numbers of pairwise constraints, which demonstrates that incorporating the augmented constraints consistently improves the clustering performance under varying numbers of queried constraints. {The advantage is more obvious when the number of constraints is large, since more entries in the similarity matrix $Y$ are labeled and the augmented constraints are more accurate. In addition, the trend that ARI increases as the number of constraints grows is more stable with the augmented constraints compared with its counterpart, indicating that the proposed method also has an increasing level of {robustness} in clustering. One possible reason is that the proposed method utilizes all $n(n-1)/2$ pairs during metric training instead of selected constraints only, which alleviates randomness and avoids overfitting labeled pairs.}

\subsection{Active clustering with low signal-noise ratio}
Another novelty of the proposed method is performing feature selection in the process of active clustering. In this simulation, we demonstrate this advantage in a low signal-noise ratio setting where the number of irrelevant features is much larger than the true features. We adopt the data generating procedure in Section \ref{sub:the_effectiveness_of_incorporating_}, except letting $P(\mu^{(2)}_{j}=c\mathbbm{1}\{j=z_i^{(2)}\})=P(\mu^{(2)}_{j}=-c\mathbbm{1}\{j=z_i^{(2)}\})=\frac{1}{2}$ to make clustering more difficult {in that the irrelevant features are well-separated}. Therefore, clustering without identifying the  true features is likely to underperform in this case.

We compare the proposed method with other popular active semi-supervised clustering methods under the setting of $p_1=5$, $p_2=30$, $c=3$, $K=5$ and $p_1=10$, $p_2=30$, $c=3$, $K=10$, respectively. In each case, we generate $n=300$ samples which are evenly distributed sampled from $K$ clusters. In addition, we let the penalty parameter $\lambda=0.5$ and the number of penalized features $q=p_2/2$. 

The competing methods include constrained Kmeans (COPKmeans) \cite{Wagstaff2001ConstrainedKC}, pairwise constrained Kmeans (PCKmeans) \cite{basu_active_2004}, metric pairwise constrained Kmeans (MPCKmeans) \cite{bilenko_integrating_2004}, constraint-based repeated aggregation (COBRA) \cite{van_craenendonck_cobra:_2018} {and constraint-based repeated aggregation and Splitting (COBRAS) \cite{duivesteijn_cobras_2018}}. Among these methods, COP-Kmeans, PCKmeans and MPCKmeans are centroid-based clustering algorithms, and COP-Kmeans and PCKmeans do not involve metric learning. The aforementioned methods are originally designed for one-time-selected pairwise constraints. To make a fair comparison, we combined these three methods with the normalized point-based uncertainty (NPU) active query strategy implemented by \cite{Jakub2018}. On the other hand, COBRA is a model-free hierarchical clustering method, by first preclustering the instances into several local neighborhoods called \textit{super-instances}, and then further combining these super-instances based on pairwise constraints. The budget of query is controlled indirectly by the initial number of super-instances. {Finally, COBRAS extends COBRA by controlling the number of queries directly and is not biased towards ellipsoidal clusters. We implement the COBRA and COBRAS with the package \cite{craenendonck2017} and \cite{cobras2020}, respectively.}

To illustrate the improvement in our clustering performance from each section separately, we break down the proposed method into two parts: the Augmented Query Metric learning method (AQM), and the Minimum Expected Entropy (MEE) active query strategy. We denote the full implementation of the proposed method as AQM+MEE. In addition, we also provide the results of combining the proposed metric learning method with the competing active strategy NPU denoted as AQM+NPU.
\color{black}

\begin{table}[H]
  \caption{Comparisons on the simulation data with $p_1=5$, $p_2=30$, $c=3$ and $K=5$, showing average ARI of clustering with standard deviations. The second half of the table provides the results of COBRA.}
  \label{tab:simulation_p1_5_p2_30}
  \centering
  \begin{adjustbox}{max width=\textwidth}
  \begin{tabular}{lccccc}
  \hline
  Number of queries & 60           & 120          & 180          & 240          & 300     \\    
  \hline
  PCKmeans + NPU            & 0.291(0.083) & 0.347(0.100) & 0.377(0.115) & 0.407(0.112) & 0.429(0.123) \\
  COPKmeans + NPU           & 0.276(0.125) & 0.369(0.147) & 0.453(0.186) & 0.552(0.198) & 0.644(0.202) \\
  MPCKmeans + NPU           & 0.275(0.115) & 0.355(0.113) & 0.434(0.131) & 0.444(0.157) & 0.529(0.157) \\
  COBRAS 					& 0.168(0.083) & 0.250(0.083) & 0.300(0.084) & 0.348(0.077) & 0.417(0.077) \\
  \color{black}
  {AQM + NPU} &  { 0.443(0.128)} &  {0.587(0.112)} & {0.688(0.090)} & {0.768(0.132)} & {0.860(0.151)} \\
  AQM + MEE  & {0.474(0.124)} &  {0.630(0.123)} & {0.725(0.147)} & {0.845(0.110)} & {0.921(0.078)} \\
 
  \hline
  \end{tabular}
  \end{adjustbox}

  \begin{adjustbox}{max width=\textwidth}
  \begin{tabular}{lccccc}
  \hline
  Number of super instances & 10            & 50              & 90              & 130             & 170             \\
  \hline
  Number of queries & 19.100(2.737) & 101.900(14.614) & 179.533(20.713) & 250.000(24.220) & 321.933(28.578) \\
  ARI of COBRA      & 0.174(0.065)  & 0.275(0.070)    & 0.365(0.065)    & 0.473(0.057)    & 0.584(0.056)    \\
  \hline
  \end{tabular}
  \end{adjustbox}
\end{table}
\begin{table}[H]
	\caption{Comparisons on the simulation data with $p_1=10$, $p_2=30$, $c=3$ and $K=10$, showing average ARI of clustering with standard deviations. The second half of the table provides the results of COBRA.}
	\label{tab:simulation_p1_10_p2_30}
	\centering
	\begin{adjustbox}{max width=\textwidth}
	\begin{tabular}{lccccc}
	\hline
	Number of queries & 60           & 120          & 180          & 240          & 300          \\
	\hline
	PCKmeans + NPU            & 0.109(0.021) & 0.131(0.034) & 0.177(0.033) & 0.200(0.049) & 0.228(0.055) \\
	COPKmeans + NPU           & 0.107(0.030) & 0.122(0.031) & 0.146(0.041) & 0.174(0.048) & 0.194(0.047) \\
	MPCKmeans + NPU           & 0.086(0.024) & 0.129(0.034) & 0.164(0.040) & 0.189(0.044) & 0.233(0.042) \\
	COBRAS 					  & 0.064(0.031) & 0.089(0.035) & 0.084(0.037) & 0.140(0.051) & 0.157(0.068) \\
	\color{black}
	AQM + NPU & 0.119(0.040)	& {0.209(0.050)}	 &{0.267(0.062)}	&0.307(0.083)	&0.350(0.082)\\
	AQM + MEE  & 	{0.131(0.049)} & {0.203(0.055)}& {0.263(0.063)} & {0.327(0.076)} & {0.350(0.090)} \\ 
	\hline
	\end{tabular}
	\end{adjustbox}
	\begin{adjustbox}{max width=\textwidth}
	\begin{tabular}{lccccc}
	\hline
	Number of super instances & 10            & 30              & 50              & 70              & 90              \\
	\hline
	Number of queries & 28.633(5.238) & 115.067(12.228) & 183.933(18.482) & 251.367(22.129) & 315.800(25.665) \\
	ARI of COBRA   & 0.076(0.025)  & 0.105(0.024)    & 0.143(0.025)    & 0.192(0.036)    & 0.242(0.031)    \\
	\hline
	\end{tabular}
	\end{adjustbox}
\end{table}

Tables \ref{tab:simulation_p1_5_p2_30} and \ref{tab:simulation_p1_10_p2_30} present the average ARI and standard deviation based on 30 replications under different numbers of queries, implying that the combination of AQM+MEE method achieves the best clustering result  when the signal-noise ratio is low, regardless of the number of queries. In particular, the proposed AQM method achieves the largest improvements on ARI by more than 50\% compared with the MPCKmeans when both methods adopt the NPU strategy. In addition, the comparison between AQM+NPU and AQM+MEE shows that the proposed active strategy MEE can further enhance clustering efficiency, especially when $p_1=5$ and $p_2=30$.  We also notice that for the methods without metric learning, namely PCKmeans, COPKmeans, COBRA and COBRAS, their performances are similar as they are not designed for extracting features through metric learning. However, although MPCKmeans involves metric learning, its accuracy is still relatively low since it does not exclude the irrelevant features. In contrast, the proposed methods improves the clustering accuracy significantly in both simulation settings, indicating the effectiveness of both AQM and MEE for clustering tasks actively.
\color{black}

	In addition, we compare the the proposed method and the competing methods under a high-dimensional setting. Specifically, we sample the relevant features from a $K$-mode Gaussian mixture model, where the $K$ centers are uniformly located on a $p_1-1$ sphere with a radius $r$. Under this setting, $K$ no longer depends on $p_1$, and the total number of features $p_1+p_2$ can be large. 

	We let $K=5,~p_1=100,~p_2=400,~r=5$ and generate $250$ samples in total. Table \ref{tab:simulation_high_dim}  provides the ARI of AQM+MEE and the competing methods with different numbers of queries. The simulation results show that the proposed method still achieves the highest clustering accuracy, despite a higher dimension of $p$. This simulation affirms that the proposed method is applicable under high dimensional cases, and can adapt to different clustering center distributions.

	\begin{table}[hbtp]
		\caption{ARI comparisons with cluster center on sphere with $K=5, p_1=100, p_2=400, r=5$ and $n=250$, showing the average ARI of clustering with standard deviations.}
		\label{tab:simulation_high_dim}
		\centering
		\begin{adjustbox}{max width=\textwidth}
		\begin{tabular}{lccccc}
		\hline
		Number of super instances & 60           & 120          & 180          & 240          & 300          \\
		\hline
		PCKmeans + NPU            & {0.391(0.051)} & 0.456(0.071) & 0.575(0.074) & 0.687(0.074) & 0.819(0.103) \\
		COPKmeans + NPU          & 0.326(0.027) & 0.374(0.041) & 0.490(0.059) & 0.562(0.067) & 0.673(0.074) \\
		MPCKmeans + NPU             & 0.325(0.094) & 0.435(0.077) & 0.564(0.060) & 0.713(0.056) & 0.824(0.049) \\
		COBRAS 						& 0.313(0.042) & 0.349(0.035) & 0.367(0.044) & 0.438(0.037) & 0.501(0.053) \\
		\color{black}
		AQM + MEE    & 0.384(0.045) &{ 0.489(0.099)} & {0.595(0.097)} & {0.773(0.037)} & {0.906(0.083)} \\ 
		\hline
		\end{tabular}
		\end{adjustbox}
		\begin{adjustbox}{max width=\textwidth}
		\begin{tabular}{lccccc}
		\hline
		Number of super instances & 30     & 70       & 90       & 110   & 150    \\
		\hline
		Number of queries & 55.400(5.389) & 141.300(5.675) & 180.200(12.983) & 222.500(12.902) & 320.000(9.571) \\
		ARI of COBRA     & 0.248(0.017)  & 0.351(0.023)   & 0.396(0.016)    & 0.454(0.015)    & 0.597(0.020)   \\
		\hline
		\end{tabular}
		\end{adjustbox}
	\end{table}  
\color{black}

\section{Real Data}
\label{sec:real_data}
We apply the proposed method on three real datasets with high dimensional features. The first dataset is the breast cancer diagnostic data from the UCI machine learning repository \cite{Dua:2019}. The dataset contains 569 samples with 30 features extracted from the diagnostic images of a breast mass, which are labeled as either malignant or benign. The second dataset is the MEU-Mobile dataset which records 71 keystroke features of phone users, including finger area, pressure and hold time, etc. We use a subset of the keystroke data from 9 users. Each user repeats typing the same password 51 times, so there are 459 samples in total. The third dataset is the urban land cover dataset which contains 675 multi-scale remote sensing images. For each resolution scale, 21 features are measured, including area, brightness, asymmetry, etc. These features are repeatedly constructed for the same image under 7 different resolutions, resulting in 147 features in total. Based on the extracted features, the images of the third dataset are categorized into 9 urban land cover classes including trees, grass, soil, concrete, asphalt, buildings, cars, pools and shadows. 

{The goal of our study is to cluster the datasets with sequentially queried pairwise constraints while identifying important features. In our implementation, we cluster the three datasets into 2, 9 and 9 categories, respectively. The category labels in the raw datasets are used only in determining the similarity of the queried instance pairs, and in verifying the accuracy of the clustering outcomes through ARI.}

\begin{figure}[btp]
  \begin{subfigure}{0.33\linewidth}
    \includegraphics[width=\linewidth]{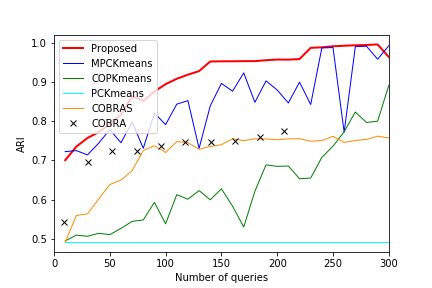}
    \caption{Breast cancer}
  \end{subfigure}
  \begin{subfigure}{0.33\linewidth}
    \includegraphics[width=\linewidth]{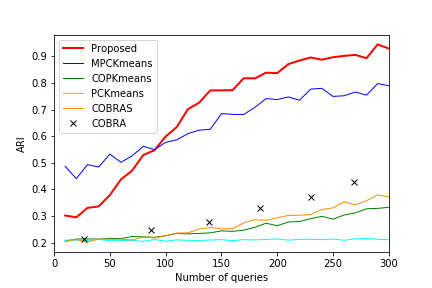}
    \caption{MEU-Mobile}
  \end{subfigure}
  \begin{subfigure}{0.33\linewidth}
    \includegraphics[width=\linewidth]{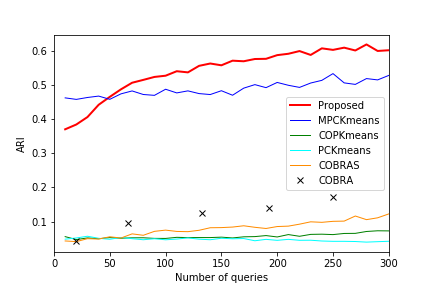}
    \caption{Urban land cover}
  \end{subfigure}
  \caption{Performance comparison on three real datasets, showing average ARI against number of constraints based on 30 replications. (The three competing methods MPCKmeans, COPKmeans and PCKmeans are combined with NPU strategy.)}
  \label{fig:real_data}
\end{figure}

{Figure~\ref{fig:real_data} provides the ARI comparison between the proposed method (AQM+MEE) and the competing methods. The proposed method has the overall best performance in all three datasets in terms of the average ARI with varying number of constraints. The most competitive method is the MPCKmeans, as it also applies metric learning to extract the important features, while the other competing approaches, i.e., PCKmeans, COPKmeans and COBRA do not. For the breast cancer data, although the proposed method reaches a similar accuracy as the MPCKmeans when the number of pairwise constraints is large, the proposed method has a higher ARI when the number of constraints is relatively small due to the higher efficiency in utilizing constraints.} In particular, the proposed method achieves an ARI of 0.928 with 80 queries while the MPCKmeans achieves only 0.732 with the same number of queries. For the MEU-Mobile data and urban land cover data, the proposed method improves the average ARI by 20\% and 17\%, respectively, compared with the MPCKmeans given 300  queries. {In addition, the ARI of MPCKmeans and COPKmeans fluctuate significantly in clustering breast cancer data as the number of queries grows. In contrast, the proposed method leads to a more stable clustering with consistently increasing ARI when the query process continues. One possible reason is that the proposed method removes the irrelevant features through selecting penalty functions. Another possible reason is that the proposed MEE strategy selects the unlabeled pairs which consistently contribute to clustering by evaluating the information gain of the selected queries more accurately compared with the NPU method.}

\begin{figure}[htbp]
  \centering
  \includegraphics[width=\linewidth]{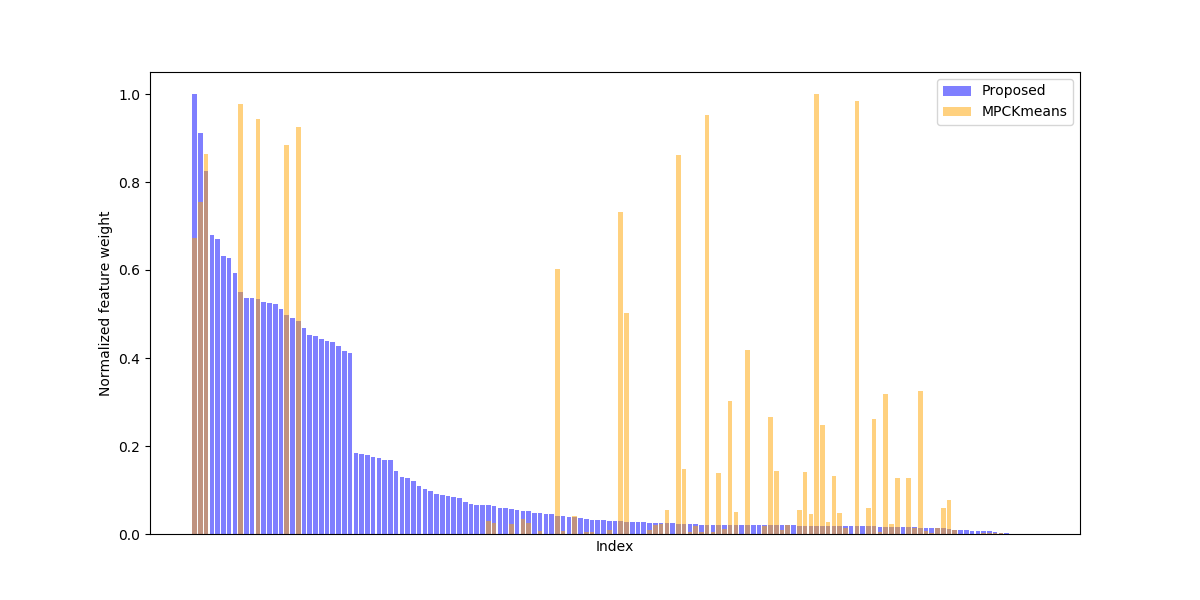}
  \caption{The estimated weights of 147 features from the urban land cover dataset, showing the significant features identified by the proposed method compared with MPCKmeans.}
  \label{fig:urban_land_cover_weight}
\end{figure}

Furthermore, we investigate the interpretability of the features selected by the proposed method via the example of the urban land cover data. We compare the significance of each feature by plotting the feature weights in a decreasing order as shown in Figure~\ref{fig:urban_land_cover_weight}. The barplot of Figure~\ref{fig:urban_land_cover_weight} implies that the proposed method divides the feature into 3 groups. The first 3 largest coefficients correspond to the 19th, 40th and 61st features in the original data. These three features are associated with the normalized difference vegetation index (NDVI) on three different resolution scales, respectively, which implies that NDVI is a crucial factor identified by the proposed method in determining the image category. The second group consists of 25 features, and the third group consists of the rest of the features which are less important in image clustering. In contrast, the MPCKmeans tends to assign high importance to the irrelevant features and the selected features are not consistent across different resolution scales.

To verify the importance of the selected features, we perform the Kmeans with the entire 147 features, the first 3 features and the 28 features in the first two groups, respectively, without imposing any pairwise constraints. With all 147 features, the Kmeans returns an ARI of 0.03. In contrast, with the first three features only, the ARI increases significantly to 0.29. With the selected 28 features, the ARI further increases to 0.34. The above results imply that the first three features extracted by the proposed method play an essential role in determining the category of the sampled remote sensing image. Although involving less important features from the second group can improve the performance, the improvement is quite negligible. On the other hand, including more features brings more irrelevant information in measuring the similarity between two images, leading to a noisy metric space with more difficulties in clustering. The real data analyses confirm that the proposed method is able to identify the low-dimensional feature space which is highly correlated to clustering analyses.
\color{black}

\section{Discussion}
\label{sec:discussion}
In this paper, we propose a novel active clustering method with metric learning. The paper has three contributions: First, we augment the queried instance-level similarity by generalizing the pairwise constraints utilizing the cluster structure that is typically ignored in the existing metric learning methods. Second, we improve the {robustness} of the metric learning process by selectively penalizing the potentially irrelevant features based on history training results. Third, we propose a new active query strategy based on the expected entropy change, which makes a more accurate evaluation of the information gain from a query. We also investigate the theoretical properties of the proposed approach, especially on the advantage between the active query strategy over random selection from the perspective of the posterior distribution of the cluster membership, which has not been studied in the existing literature to the best of our knowledge. Finally, we demonstrate the efficacy of the proposed method through simulation studies and real data applications in breast cancer diagnosis, keystroke recognition and multi-scale remote sensing images.

The proposed framework can be extended to online training in that both constraint augmentation and metric aggregation can be adapted into a progressive method without retraining at each step when new constraints are added, which can improve computation efficiency. {In addition, the proposed method can be generalized to fit the non-ellipsoidal clusters through non-linear feature transformation, such as the kernel method  \cite{anandSemiSupervisedKernelMean2014a,abinActiveConstrainedFuzzy2015}, or adopting other constrained clustering methods instead of PCKmeans, such as spectral clustering methods \cite{wangActiveSpectralClustering2010,hsin-chienhuangAffinityAggregationSpectral2012} (See Supplementary Materials for details).} On the other hand, the theoretical properties indicate that metric learning and active query can be interpreted as optimizing the likelihood function and the prior function sequentially regarding cluster membership distribution, respectively. Therefore, one future research direction is to develop a unified framework by incorporating the randomness of metric learning into the active query process to enhance {robustness}. Another potential research direction is to extend the current method under the setting of the model-free constraints generating process using deep learning tools such as Generative Adversarial Networks (GAN) \cite{goodfellow2014generative}.
\color{black}

\section*{Supplementary Materials}
The online supplement contains all technical proofs and additional numerical results.

\section*{Acknowledgment}
The authors are grateful to reviewers, the Associate Editor and Editor for their insightful comments and suggestions which have improved the manuscript significantly.
{\footnotesize
\bibliography{JASA_revision2}
}
\end{document}